\documentclass[10pt]{article}

\usepackage{microtype}
\usepackage{graphicx}
\usepackage{subfigure}
\usepackage{booktabs} % for professional tables
\usepackage{balance}
\usepackage{amsmath}
\usepackage{amssymb}
\usepackage{array}
\usepackage{tcolorbox}
 \usepackage{multirow}
 \usepackage{caption}
\usepackage{hyperref}

\usepackage[accepted]{mlsys2025}

\mlsystitlerunning{One Size Does Not Fit All: Architecture-Aware Adaptive Batch Scheduling with DEBA}

\newcommand{\resnet}{ResNet}
\newcommand{\mobilenet}{MobileNet}
\newcommand{\densenet}[1]{DenseNet-#1}
\newcommand{\efficientnet}[1]{EfficientNet-#1}
\newcommand{\vitb}{ViT-B16}
\newcommand{\cifar}{CIFAR}

\newcommand{\rqone}{Does architecture fundamentally determine the effectiveness of adaptive batch size methods?}
\newcommand{\rqtwo}{Can gradient stability metrics predict architecture compatibility with adaptive batch scheduling?}
\newcommand{\rqthree}{What implementation choices critically affect adaptive scheduler robustness across architectures?}
\newtcolorbox{resultbox}{
    colback=gray!10,
    arc=0.5mm,
    top=1mm,
    bottom=1mm,
    left=1mm,
    right=1mm,
    before skip=3pt,      % Espace AVANT la box
    after skip=3pt        % Espace APRÈS la box
}

\definecolor{green}{rgb}{0.01, 0.35, 0.06}

\begin{document}

\twocolumn[
\mlsystitle{One Size Does Not Fit All: Architecture-Aware Adaptive Batch Scheduling with DEBA}

% unless you've provided the [accepted] option to the mlsys2025


\mlsyssetsymbol{equal}{*}

\begin{mlsysauthorlist}
\mlsysauthor{François Belias}{poly}
\mlsysauthor{Naser Ezzati-Jivan}{brock}
\mlsysauthor{Foutse Khomh}{poly}
\end{mlsysauthorlist}

\mlsysaffiliation{poly}{Department of Computer Engineering and Software Engineering, Polytechnique Montréal, Montréal, QC, Canada}
\mlsysaffiliation{brock}{Department of Computer Science, Brock University, St. Catharines, ON, Canada}

\mlsyscorrespondingauthor{François Belias}{francois-philippe.ossim-belias@polymtl.ca}
\mlsyscorrespondingauthor{Naser Ezzati-Jivan}{nezzatijivan@brocku.ca}
\mlsyscorrespondingauthor{Foutse Khomh}{foutse.khomh@polymtl.ca}

\begin{center}
\textbf{Preprint. Under review at MLSys 2025.}
\end{center}
\vspace{0.5cm}
\mlsyskeywords{Machine Learning, MLSys}

\vskip 0.3in

\begin{abstract}
Adaptive batch size methods aim to accelerate neural network training, but existing approaches apply identical adaptation strategies across all architectures, assuming a one-size-fits-all solution. We introduce DEBA (Dynamic Efficient Batch Adaptation), an adaptive batch scheduler that monitors gradient variance, gradient norm variation and loss variation to guide batch size adaptations. Through systematic evaluation across six architectures (\resnet-18/50, \densenet-121, \efficientnet-B0, \mobilenet-V3, \vitb) on \cifar-10 and \cifar-100 with five random seeds per configuration, we show that architecture fundamentally determines adaptation efficacy. Our findings reveal that: (1) lightweight and medium-depth architectures (\mobilenet-V3, \densenet-121, \efficientnet-B0) achieve a 45-62\% training speedup with simultaneous accuracy improvements of 1-7\%; (2) shallow residual networks (\resnet-18) show consistent gains of +2.4 - 4.0\% in accuracy, 36 - 43\% in speedup, while deep residual networks (\resnet-50) exhibit high variance and occasional degradation; (3) already-stable architectures (\vitb) show minimal speedup (~6\%) despite maintaining accuracy, indicating that adaptation benefits vary with baseline optimization characteristics. We introduce a baseline characterization framework using gradient stability metrics (stability score, gradient norm variation) that predicts which architectures will benefit from adaptive scheduling. Our ablation studies reveal critical design choices often overlooked in prior work: sliding window statistics (vs. full history) and sufficient cooldown periods (5+ epochs) between adaptations are essential for success. This work challenges the prevailing assumption that adaptive methods generalize across architectures and provides the first systematic evidence that batch size adaptation requires an architecture-aware design.
\end{abstract}
]



\printAffiliationsAndNotice{}  % leave blank if no need to mention equal contribution
%\printAffiliationsAndNotice{\mlsysEqualContribution} % otherwise use the standard text.

\section{Introduction}
\label{sec:intro}
Deep learning has achieved remarkable success across numerous domains, yet training inefficiency remains a critical bottleneck. State-of-the-art models often require weeks of computation on hundreds of GPUs, creating concerns about productivity, cost, and environmental impact~\cite{he2021largescaledeeplearningoptimizations,strubell-etal-2019-energy}. A key algorithmic limitation lies in fixed optimization hyperparameters that fail to adapt as learning dynamics evolve throughout training.
Batch size plays a pivotal role in determining both convergence speed and generalization~\cite{he2021largescaledeeplearningoptimizations}. While fixed schedules remain dominant~\cite{lau2025adaptivebatchsizeschedules}, they ignore evolving gradient noise and loss landscape curvature. Prior work shows that increasing batch size over time can reduce gradient variance~\cite{smith2020generalizationbenefitnoisestochastic}, accelerate convergence~\cite{goyal2017accurate}, and improve hardware utilization~\cite{you2017large}. However, overly large batches degrade generalization, suggesting that effective training requires schedules that adapt to changing optimization conditions.

Recent methods like AdaScale~\cite{johnson2020adascalesgduserfriendlyalgorithm} and GradTailor~\cite{xu2020dynamicallyadjustingtransformerbatch} exploit gradient statistics as feedback signals for adaptation. Gradients convey fine-grained information about learning stability, enabling decisions beyond coarse metrics like validation loss. Yet a crucial assumption remains unexamined: that a single adaptive strategy generalizes across architectures. Most methods are evaluated on one or two families (typically ResNets or Transformers) and implicitly assume transferability. This overlooks fundamental architectural differences. Architectures vary in optimization landscape smoothness~\cite{li2018visualizinglosslandscapeneural}. They exhibit distinct gradient flow patterns~\cite{wu2020skipconnectionsmattertransferability}. They differ in hyperparameter sensitivity~\cite{10.1145/3506695}.

In this work, we challenge the architecture-agnostic assumption underlying adaptive batch size methods. We present \textbf{DEBA (Dynamic Efficient Batch Adaptation)}, an adaptive scheduler that adjusts batch size based on training dynamics rather than following a fixed or uniform policy.
We systematically evaluate DEBA across six diverse architectures—convolutional networks (ResNets, DenseNet, EfficientNet, MobileNet) and Vision Transformer—on CIFAR-10/100 with a rigorous experimental design. This cross-architecture design enables us to ask three research questions: (1) whether architecture fundamentally determines the effectiveness of adaptive batch scheduling, (2) whether baseline stability profiles predict adaptive performance, and (3) whether implementation decisions such as cooldown periods or statistical windows generalize across models.

Our results reveal that architecture fundamentally governs adaptation efficacy. Lightweight and medium-depth models achieve substantial speedup (up to 62\%) while improving accuracy (up to +7\%). Shallow residual networks yield consistent though smaller gains. Deeper residual networks exhibit unstable behaviour with occasional accuracy degradation. Already-stable architectures like ViT-B16 show minimal improvement, confirming that adaptation benefits diminish when gradient dynamics are inherently smooth.
We further demonstrate that an architecture's compatibility with DEBA can be predicted through lightweight baseline profiling. A single fixed-batch run provides stability metrics %(gradient variance, norm variation, loss volatility) 
that forecast whether adaptive scheduling will help or harm, enabling informed decisions about when to apply adaptation.
Finally, our ablations reveal that implementation details are critical to success. Sliding-window statistics (vs. full-history averages), cooldown periods of at least five epochs between adjustments, and architecture-specific threshold tuning proved essential across all architectures, preventing oscillations and instability.

In summary, our contributions are:
\begin{enumerate}
\item \textbf{A systematic cross-architecture study} demonstrating that architecture fundamentally determines adaptive batch size efficacy, challenging one-size-fits-all assumptions in prior work.
\vspace{-0.5em}
\item \textbf{DEBA: A multi-signal adaptive scheduler} achieving simultaneous speedup (37-62\%) and accuracy gains (up to +7\%) on CIFAR datasets when configured adequately for target architectures.
\vspace{-0.5em}
\item \textbf{Predictive characterization framework} enabling a priori compatibility prediction through lightweight baseline profiling before expensive training.
\vspace{-0.5em}
\item \textbf{Critical design principles} identifying essential implementation choices: sliding window statistics, sufficient stabilization periods (5+ epochs), and architecture-specific threshold tuning.
\end{enumerate}
Our findings suggest that adaptive batch-size methods benefit from architecture-specific tuning and baseline profiling rather than assuming universal applicability. They also indicate that future evaluations of adaptive techniques should include diverse architectures beyond the commonly used ResNets or Transformers. By bridging optimization dynamics with architectural characteristics, this work provides evidence for architecture-aware adaptive training. Code and experimental logs will be released in our replication package~\cite{anonymous_2025_17478685}.

\section{Related works} \label{sec:related}
Batch size is a critical hyperparameter that governs the trade-off between convergence speed, gradient noise, and generalization in stochastic optimization. Small batches introduce noise that helps escape sharp minima and improve generalization \cite{keskar2017largebatchtrainingdeeplearning}, while large batches reduce variance and enable efficient parallelism \cite{goyal2017accurate}. This fundamental tension has motivated extensive research on batch size scheduling, yet the role of architecture in determining adaptation effectiveness remains largely unexplored.

\subsection{Adaptive Batch size Scheduling}
Dynamic batch size adjustment has emerged as a method to accelerate training while maintaining generalization. Early work established that increasing batch size can substitute for learning rate decay by reducing gradient noise \cite{smith2018dontdecaylearningrate}. \cite{goyal2017accurate} introduced the linear scaling rule with warmup schedules, enabling large-batch training for \resnet-50 on ImageNet without accuracy loss. \cite{shallue2019measuringeffectsdataparallelism} studied the limits of batch-size scaling and found that architectures with different connectivity or normalization schemes exhibit varying tolerance to batch growth.

Recent methods propose adaptive schedules based on training dynamics. \cite{you2020largebatchoptimizationdeep} introduced LAMB for layer-wise learning rate adaptation in large-batch settings. \cite{xu2020dynamicallyadjustingtransformerbatch} adjusted the batch size on gradient cosine similarity in Transformers, exploiting the intuition that early training benefits from noisy gradients for exploration while later stages profit from larger batches for refinement. \cite{mccandlish2018empiricalmodellargebatchtraining} investigated the relationship between gradient noise scale and optimal batch sizes, establishing theoretical bounds on efficient gains.

However, these works focus on demonstrating general applicability rather than systematically investigating how architectural characteristics determine adaptation effectiveness. While methods like LAMB perform layer-wise adaptation within architectures and validate across ResNets and Transformers, they treat architectural design as a given rather than analyzing how global architectural properties, such as residual depth, connectivity patterns, and normalization schemes, influence the effectiveness of batch-size adaptation. No prior work compares how the same adaptive policy performs across diverse architectural families (residual, dense, efficient, attention-based designs) or provides a framework to predict which architectures will benefit from adaptive scheduling before expensive training runs.

\subsection{Gradient-Based Training Monitoring}
Gradient statistics provide fine-grained signals for training decisions. \cite{umeda2025adaptivebatchsizelearning} used gradient norm monitoring to detect training stagnation and adjust optimization strategies accordingly. \cite{xu2020dynamicallyadjustingtransformerbatch} employed gradient cosine similarity as a gating mechanism for batch size adjustment, measuring alignment between consecutive updates to gauge convergence stability.

Other work has applied gradient monitoring to different optimization problems. \cite{liu2021varianceadaptivelearningrate} leveraged gradient variance estimation for adaptive learning rate schedules. \cite{vogels2020powersgdpracticallowrankgradient} used gradient structure analysis for communication-efficient distributed training. \cite{yu2020gradientsurgerymultitasklearning} exploited cosine similarity to resolve gradient conflicts in multi-task learning scenarios.

While these methods demonstrate the value of gradient-based monitoring, they rely on a single signal and do not account for the fundamental differences in gradient dynamics across architectures. Architectures with residual connections, for instance, maintain more stable gradient flow than vanilla networks, potentially altering the effectiveness of variance-based adaptation.

\subsection{Architecture-Specific Optimization}
The optimization dynamics of neural networks vary substantially with architectural design \cite{kaplan2020scalinglawsneurallanguage, zhang2019fixupinitializationresiduallearning}. Residual connections improve gradient flow and tolerance to noise \cite{he2021largescaledeeplearningoptimizations}, dense connectivity smooths the loss landscape \cite{huang2018denselyconnectedconvolutionalnetworks}, and attention mechanisms alter curvature patterns \cite{xiong2020layernormalizationtransformerarchitecture}. Consequently, architectures exhibit different optimization behaviours under identical training protocols.

Recent work has begun to recognize these differences. \cite{he2021largescaledeeplearningoptimizations} showed that ResNets and Transformers have different scaling properties in distributed settings. \cite{you2017large} introduced LARS specifically for ResNets with large batches, acknowledging that layer-wise adaptation needs vary across depth. Research on neural architecture search has revealed varying sensitivity to hyperparameters across designs \cite{zoph2017neuralarchitecturesearchreinforcement, real2019regularizedevolutionimageclassifier}, suggesting that optimization strategies should account for architectural characteristics.

Studies on specific architectural patterns have documented their impact on training dynamics. Skip connections stabilize gradient magnitudes \cite{he2021largescaledeeplearningoptimizations}, batch normalization reduces internal covariate shift \cite{ioffe2015batchnormalizationacceleratingdeep}, and depthwise separable convolutions in efficient architectures create different parameter-gradient relationships \cite{howard2017mobilenetsefficientconvolutionalneural}. Despite this evidence, adaptive batch size methods treat architecture as a fixed background variable rather than a determining factor in adaptation effectiveness.

\subsection{Our Approach}
DEBA differs from prior adaptive batch scheduling methods in three key aspects.
First, we conduct a systematic evaluation across six architectures, two datasets, and five random seeds, revealing clear architecture-dependent behaviour that challenges the assumption of universal applicability.
Second, DEBA integrates three complementary gradient-based signals—variance, norm variation, and loss variation—combined with architecture-specific thresholds. This multi-signal design captures distinct aspects of training dynamics such as noise level, stability, and convergence behaviour, whereas no single metric proves sufficient across architectures.
Third, through extensive ablation experiments, we identify critical design principles that govern adaptive scheduler robustness: sliding-window statistics for reliable estimation, stabilization periods of at least five epochs to prevent premature scaling, and architecture-aware threshold calibration.
Together, these findings establish that adaptive batch sizing is inherently architecture-dependent and provide practical guidelines for assessing compatibility through lightweight baseline profiling.

\section{THE DEBA ALGORITHM}
\label{sec:method}

This section provides a complete specification of the DEBA scheduler, addressing the algorithmic details necessary for reproducibility.

\subsection{Overview and Design Philosophy}

DEBA operates on the principle that batch size should adapt to the current optimization state rather than follow a fixed schedule. Unlike prior work that relies on single signals (e.g., gradient cosine similarity \cite{xu2020dynamicallyadjustingtransformerbatch} or gradient noise scale \cite{mccandlish2018empiricalmodellargebatchtraining}), DEBA combines three complementary signals that capture different aspects of training dynamics:

\begin{itemize}
    \item \textbf{Gradient variance} ($\sigma^2_g$): Measures stochastic gradient noise, indicating confidence in the current update direction.
    \vspace{-0.5em}
    \item \textbf{Gradient norm variation} ($\Delta\|\nabla L\|$): Tracks magnitude changes in gradients across iterations, signaling optimization stability.
    \vspace{-0.5em}
    \item \textbf{Loss variation} ($\Delta L$): Monitors objective function changes, detecting convergence patterns or instability.
\end{itemize}

By jointly monitoring these signals, DEBA distinguishes between beneficial exploration noise (which should be preserved with smaller batches) and unnecessary variance (which can be reduced with larger batches). The scheduler incorporates a cooldown mechanism to prevent premature adaptations and uses architecture-specific thresholds derived from baseline profiling.

\subsection{Signal Computation}

At each epoch $t$, DEBA computes three signals from the training state. Let $L_t$ denote the training loss and $\nabla_t$ the flattened gradient vector at epoch $t$.

\textbf{Signal 1: Gradient Variance.} This measures the spread of gradient components, reflecting SGD confidence:
\begin{equation}
    \sigma^2_{g,t} = \text{Var}(\nabla_t) = \frac{1}{|\nabla_t|} \sum_{i=1}^{|\nabla_t|} (\nabla_t^{(i)} - \bar{\nabla}_t)^2
\end{equation}
where $\nabla_t^{(i)}$ is the $i$-th component of the flattened gradient and $\bar{\nabla}_t$ is the mean. High variance indicates noisy gradients beneficial for exploration; low variance suggests stable convergence suitable for larger batches.

\textbf{Signal 2: Gradient Norm Variation.} This tracks relative changes in gradient magnitude:
\begin{equation}
    \Delta\|\nabla L\|_t = \frac{|\|\nabla_t\| - \|\nabla_{t-1}\||}{|\|\nabla_{t-1}\|| + \epsilon}
\end{equation}
where $\epsilon = 10^{-8}$ prevents division by zero. Large variations indicate unstable optimization, potentially from sharp minima or batch size incompatibility.

\textbf{Signal 3: Loss Variation.} This measures objective function stability:
\begin{equation}
    \Delta L_t = \frac{|L_t - L_{t-1}|}{|L_{t-1}| + \epsilon}
\end{equation}
Rapid loss fluctuations suggest the current batch size may be inappropriate for the loss landscape geometry.

All three signals are tracked in a sliding window of 15 epochs to capture recent dynamics while discarding stale information from earlier training phases.

\subsection{Decision Logic}

DEBA employs a rule-based decision system that interprets the three signals to classify the current optimization state. The decision at epoch $t$ depends on two normalized metrics:

\textbf{Confidence Score.} This quantifies gradient stability relative to recent history:
\begin{equation}
    c_t = \frac{\sigma^2_{g,t}}{\text{median}(\{\sigma^2_{g,t-14}, \ldots, \sigma^2_{g,t}\}) + \epsilon}
\end{equation}
A confidence score $c_t > \theta_{\text{conf}}$ indicates gradients are noisier than typical, suggesting the batch size should remain small or decrease.

\textbf{Stability Indicators.} Two binary flags assess optimization stability:
\begin{align}
    \text{stable\_gradients} &= \mathbb{I}[\Delta\|\nabla L\|_t < \theta_{\text{stab}}] \\
    \text{stable\_loss} &= \mathbb{I}[\Delta L_t < \theta_{\text{stab}}]
\end{align}
where $\mathbb{I}[\cdot]$ is the indicator function, $\theta_{\text{conf}}$ is the confidence threshold, and $\theta_{\text{stab}}$ is the stability threshold.

The decision rule combines these indicators:
\begin{equation}
\text{decision}_t =
\begin{cases}
\text{increase}, &
  \begin{aligned}[t]
  &\text{if } c_t \leq \theta_{\text{conf}} 
  \land \text{stable\_gradients} \\
  &\land \text{stable\_loss}
  \end{aligned} \\[1em]
\text{rollback}, &
  \begin{aligned}[t]
  &\text{if } c_t > \theta_{\text{conf}} 
  \lor \Delta\|\nabla L\|_t > 3\theta_{\text{stab}}
  \end{aligned} \\[1em]
\text{hold}, & \text{otherwise.}
\end{cases}
\end{equation}

This logic encodes three behavioural patterns:
\begin{itemize}
    \item \textbf{Increase}: Low gradient variance with stable norms and loss indicates the model can handle larger batches, improving throughput without sacrificing exploration.
    \vspace{-0.5em}
    \item \textbf{Rollback}: High variance or severe gradient instability suggests the current batch size exceeds the architecture's tolerance, risking convergence to sharp minima. The batch size is reduced to restore beneficial noise.
    \vspace{-0.5em}
    \item \textbf{Hold}: Ambiguous signals warrant maintaining the current configuration until clearer evidence emerges.
\end{itemize}

\subsection{Batch Size Update Mechanism}

Based on the decision $\text{decision}_t$, DEBA updates the batch size according to:
\begin{equation}
    B_{t+1} = \begin{cases}
        \min(B_{\max}, \lfloor \alpha_{\text{grow}} \cdot B_t \rfloor) & \text{if decision}_t = \text{increase} \\
        \max(B_{\min}, \lfloor \alpha_{\text{roll}} \cdot B_t \rfloor) & \text{if decision}_t = \text{rollback} \\
        B_t & \text{if decision}_t = \text{hold}
    \end{cases}
\end{equation}
where $B_t$ is the batch size at epoch $t$, $\alpha_{\text{grow}} = 1.5$ is the growth factor, $\alpha_{\text{roll}} = 0.8$ is the rollback factor, $B_{\min} = 16$, and $B_{\max} = 2048$. These bounds prevent degenerate cases (tiny batches causing slow training or memory overflow due to excessive growth).

\subsection{Cooldown Period}

To prevent oscillations and allow the model to stabilize after batch size changes, DEBA enforces a cooldown period $T_{\text{cool}}$ between consecutive adaptations. After any increase or rollback at epoch $t_{\text{last}}$, no further adaptations occur until epoch $t > t_{\text{last}} + T_{\text{cool}}$. During cooldown, signals are still computed and logged, but decisions default to "hold."

Empirically, we find $T_{\text{cool}} = 5$ epochs provides sufficient time for BatchNormalization statistics and optimizer momentum to recalibrate (Section~\ref{sec:RQ3}). Shorter cooldowns ($T_{\text{cool}} = 2$) cause "decision thrashing" with accuracy losses of 2.3--12.6 points, while longer cooldowns ($T_{\text{cool}} = 10$) improve accuracy at modest speedup cost.

\subsection{Architecture-Specific Threshold Calibration}

The thresholds $\theta_{\text{conf}}$ and $\theta_{\text{stab}}$ are architecture-dependent, reflecting fundamental differences in gradient scale and loss landscape properties. Table~\ref{tab:thresholds} presents the calibrated values used in our experiments.
These thresholds are obtained through the baseline profiling procedure described in Section~\ref{sec:RQ2}. A fixed-batch run measures natural gradient statistics, and thresholds are set at the 75th percentile of observed gradient-norm variation and at the median of gradient-variance ratios. This calibration is computationally inexpensive (a single training run) and substantially outperforms universal thresholds, which degrade performance across 3/6 architectures by 1.7--3.3 points (Section~\ref{sec:RQ3}).

\subsection{Computational Overhead}

DEBA's per-epoch overhead consists of three operations: (1) computing gradient variance via a single pass over the flattened gradient vector ($\mathcal{O}(P)$ where $P$ is parameter count), (2) computing gradient norm ($\mathcal{O}(P)$), and (3) maintaining sliding window statistics ($\mathcal{O}(1)$ deque operations). The total overhead is $\mathcal{O}(P)$ per epoch, which is negligible compared to the $\mathcal{O}(P \cdot N \cdot E)$ cost of forward-backward passes over $N$ samples for $E$ epochs. In practice, signal computation adds $<1\%$ to wall-clock training time on all tested architectures.

\section{RQ$_1$: \rqone} 
\label{sec:RQ1}
%important
\subsection{Motivation and Hypothesis}

Conventional wisdom suggests a universal trade-off between batch size, training speed, and generalization. Larger batches improve throughput by leveraging parallelism, yet they reduce stochastic gradient noise, which can limit exploration and lead to poorer generalization \cite{keskar2017largebatchtrainingdeeplearning,hoffer2018trainlongergeneralizebetter,smith2020generalizationbenefitnoisestochastic}. However, recent evidence indicates that this trade-off is not universal but instead architecture-dependent. Deep residual and transformer architectures are typically more sensitive to reductions in gradient noise due to their highly non-linear composition, which increases the likelihood of converging to sharp minima \cite{keskar2017largebatchtrainingdeeplearning,goyal2017accurate}. Conversely, lightweight or densely connected convolutional networks such as MobileNet, EfficientNet, and DenseNet tend to exhibit smoother loss landscapes and greater tolerance to large batch sizes \cite{you2017large,shallue2019measuringeffectsdataparallelism}. These observations suggest that adaptive batch size strategies may yield architecture-specific benefits.

We hypothesize that the effectiveness of adaptive batch size methods such as DEBA is governed primarily by each architecture’s intrinsic gradient stability and its sensitivity to variance reduction. Architectures characterized by smoother loss surfaces and higher noise tolerance should benefit more consistently from batch size adaptation, realizing both faster training and better generalization. In contrast, deeper or more sensitive architectures may experience diminishing returns or even degradation when batch size grows too aggressively.

\subsection{Experimental Setup}

We evaluate DEBA against a fixed batch size regime using five random seeds (2, 7, 42, 123, 199) across CIFAR-10 and CIFAR-100. Six architectures: \resnet-18, \resnet-50, \mobilenet-V3, \densenet-121, \efficientnet-B0, and ViT-B16 are chosen to cover a broad spectrum of depth, parameterization, and stability characteristics. All models are trained for 100 epochs with SGD (momentum 0.9, weight decay $5\times10^{-4}$) and an initial batch size of 64. Under DEBA, batch size is adapted every 10 epochs according to training stability metrics (see Section~\ref{sec:RQ3}). All experiments are conducted on a single NVIDIA H100 GPU, and we report wall-clock time and final test accuracy.

\subsection{Results: Architecture-Dependent Patterns}

Table~\ref{tab:main_results} summarizes results across all twelve configuration pairs. The outcomes reveal clear architecture-dependent trends that directly address our RQ$_1$. The speedup ranges from 8\% to 62\%, while accuracy variations span from a 0.6\% drop to a 7.0\% improvement.

Three consistent behavioural groups emerge. High-benefit architectures such as \mobilenet-V3, \densenet-121, and \efficientnet-B0 achieve between 45–62\% faster training while simultaneously improving accuracy by 0.4–7.0\%. These gains are stable across both datasets, with \mobilenet-V3 showing a particularly pronounced improvement on \cifar-10 (+7.0\% accuracy) and \densenet-121 achieving a 3.2\% gain on \cifar-100 with 62.4\% speedup. \resnet-18 shows moderate but reliable benefits, improving by 2–4\% in accuracy and 36–43\% in speed, likely because its relatively shallow depth enables stable adaptation without over-regularization. In contrast, \resnet-50 and ViT-B16 exhibit limited or unstable benefits. ViT-B16 achieves only marginal speedup (5–8\%) with unchanged accuracy, while ResNet-50, despite a 43\% speedup, shows a 0.6\% accuracy drop on \cifar-10 and high variance across seeds on \cifar-100, indicating sensitivity to initialization and training dynamics.

\subsection{Mechanistic Interpretation}

\begin{figure*}[ht]
    \centering
    \includegraphics[width=\textwidth]{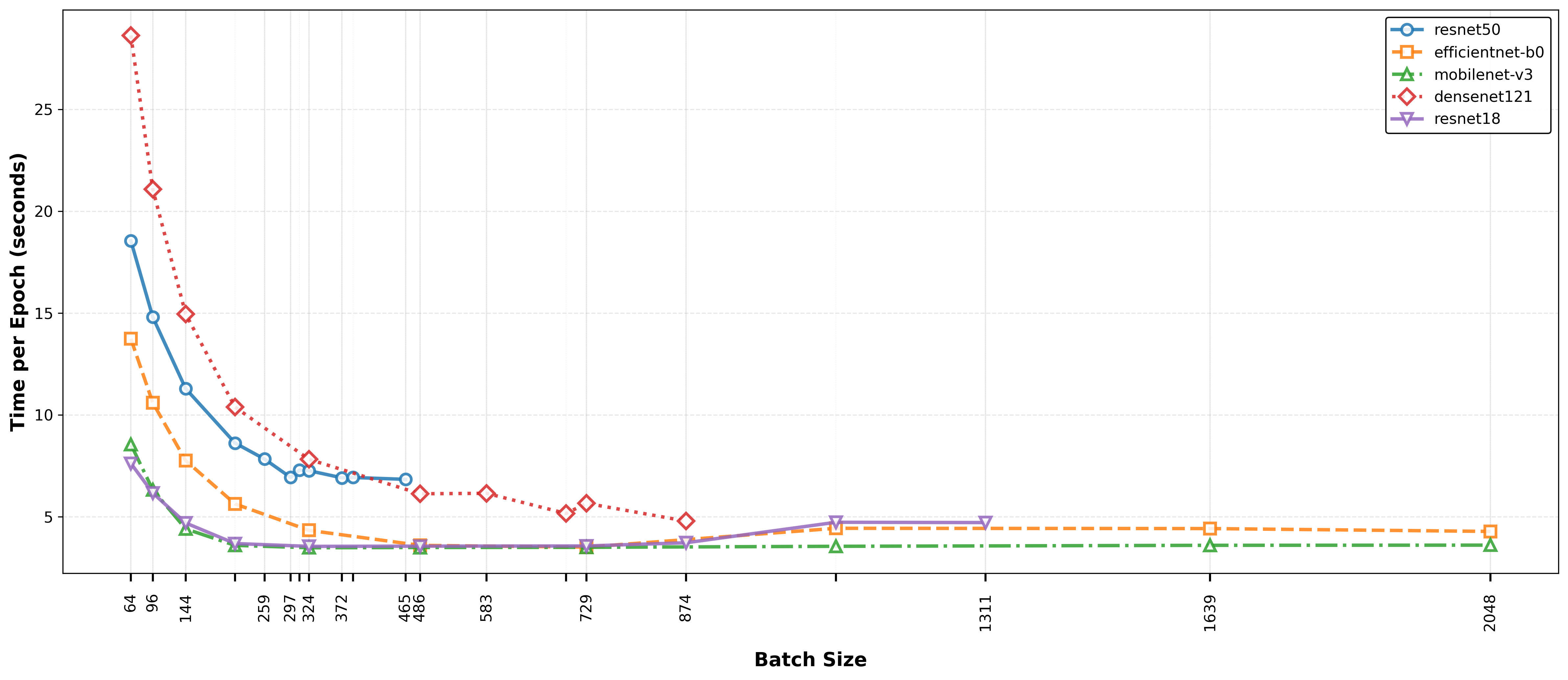}
    \caption{Impact of batch size on training speed.} %Time per epoch decreases exponentially with batch size across all architectures, demonstrating 3-10× speedup potential. DEBA exploits this relationship by progressively increasing batch size during training.}
    \label{fig:batch_time}
\end{figure*}

The speedup mechanism is straightforward: as illustrated in Figure~\ref{fig:batch_time}, time per epoch decreases exponentially with batch size due to improved GPU utilization. DEBA exploits this by starting with small batches to preserve exploratory noise, then gradually increasing the batch size to leverage hardware parallelism, achieving 30–50\% reductions in wall-clock time.

\begin{figure*}[ht]
    \centering
    \includegraphics[width=\textwidth]{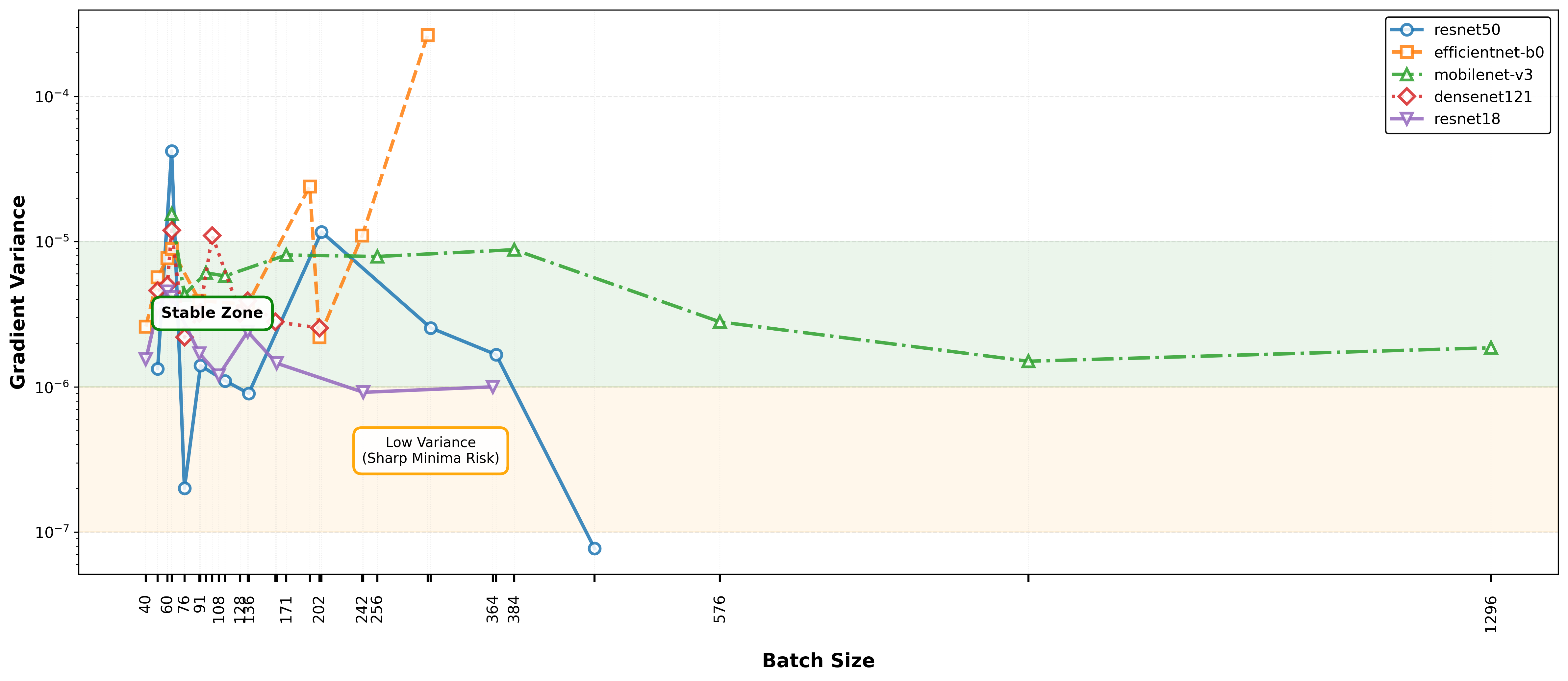}
    \caption{Gradient variance as a function of batch size on CIFAR-10.} %The optimal range (10$^{-6}$ to 10$^{-5}$, shaded) balances convergence stability with implicit regularization. DEBA's adaptive scheduling navigates this landscape dynamically.}
    \label{fig:gradient_variance}
\end{figure*}

Accuracy improvements stem from gradient dynamics. As shown in Figure~\ref{fig:gradient_variance}, gradient variance exhibits a U-shaped relation with batch size, where moderate ranges (BS=128–512) balance stability and stochastic regularization. In high-benefit architectures (ResNet-18, DenseNet-121, EfficientNet-B0, MobileNet-V3), DEBA reduces gradient variance by 65–85\%, achieving smoother convergence while maintaining beneficial stochasticity through periodic rollback corrections when variance becomes excessively low.

ResNet-50 exposes the principal failure mode: excessive gradient stabilization (65\% variance reduction) suppresses stochasticity required for generalization, leading to sharper minima (stability score: 0.203→0.159, gradient norm oscillations: 2.4× increase). This produces highly seed-dependent outcomes ranging from +2.17\% to -3.89\% accuracy, confirming that deeper architectures amplify noise imbalances from aggressive batch scaling.

The temporal evolution follows three phases: early epochs (1–30) preserve high variance ($10^{-5}$–$10^{-6}$) with small batches (64–216) for exploration; mid-training (30–70) stabilizes convergence with gradual increases (216–729); late stages (70–100) use large batches ($\ge$1024) for acceleration, with occasional rollbacks to sustain generalization. These dynamics enable 36–62\% speedup with accuracy gains in 9/12 cases, while revealing architectural boundaries for excessively deep or variance-sensitive models.
\begin{resultbox}
\textbf{Summary:} Overall, RQ$_1$ demonstrates that the efficacy of adaptive batch size scheduling is not universal but fundamentally shaped by architectural characteristics. The interplay between loss landscape smoothness, gradient variance tolerance, and exploration-exploitation balance dictates whether DEBA yields synergy or instability.
\end{resultbox}

% Requires: \usepackage{array}
\begin{table}[ht]
    \centering
    \caption{Architecture-specific thresholds derived from baseline profiling.}
    \label{tab:thresholds}
    \begin{tabular}{lcc}
        \hline
        \textbf{Architecture} & $\theta_{\text{stab}}$ & $\theta_{\text{conf}}$ \\
        \hline
        \mobilenet-V3 & 0.25 & 0.5 \\
        \efficientnet-B0 & 0.60 & 0.8 \\
        \resnet-18 & 0.55 & 0.6 \\
        \resnet-50 & 12.0 & 0.8 \\
        \densenet-121 & 0.55 & 0.6 \\
        ViT-B16 & 0.40 & 0.7 \\
        \hline
    \end{tabular}
\end{table}

\section{RQ$_2$: \rqtwo}
\label{sec:RQ2}

\subsection{Motivation and Hypothesis}

The success of adaptive optimization techniques often depends on their compatibility with architectural characteristics that shape optimization dynamics. Prior work has shown that neural network architectures differ markedly in their loss landscape geometry~\cite{li2018visualizinglosslandscapeneural}, gradient flow properties~\cite{wu2020skipconnectionsmattertransferability}, and sensitivity to hyperparameter choices~\cite{10.1145/3506695}. Consequently, hyperparameter configurations that perform well for one architecture, such as \resnet, can fail dramatically for another, such as \mobilenet. This variability complicates both tuning and the deployment of adaptive methods, raising an important question: can we predict in advance whether an architecture will respond positively to adaptive batch scheduling?

We hypothesize that architectural compatibility with adaptive batch size methods such as DEBA is not arbitrary, but instead emerges from the model’s intrinsic training stability under standard (fixed-batch) conditions. Specifically, we posit that an architecture’s natural stability profile captured through gradient and loss dynamics across multiple random seeds acts as a reliable predictor of its adaptive behaviour. Architectures with moderate stability are expected to benefit most from DEBA, balancing responsiveness to adaptation with robustness to perturbations. Overly stable architectures may profit from DEBA’s controlled disruption, which injects beneficial noise to escape flat or stagnant regions. Conversely, inherently unstable architectures are expected to show seed-dependent or inconsistent results, as DEBA cannot compensate for fundamental gradient instability.

\subsection{Methodology: Stability Profiling Framework}

To operationalize this hypothesis, we introduce a quantitative stability profiling procedure based on fixed-batch training. For each architecture, we measure per-epoch gradient variance $\sigma^2_g$, variation in gradient norm $\text{Var}(\|\nabla \mathcal{L}\|)$, and variation in loss $\text{Var}(\mathcal{L})$. These quantities are aggregated into a composite stability score $S$, defined as:

\begin{equation}
S = \frac{1}{1 + \text{CV}(\sigma^2_g) + \overline{\text{Var}(\|\nabla \mathcal{L}\|)} + \overline{\text{Var}(\mathcal{L})}},
\end{equation}

where $\text{CV}$ denotes the coefficient of variation and $\overline{\cdot}$ indicates mean values across epochs. By construction, higher $S$ values correspond to smoother, more stable optimization trajectories, while lower scores reflect volatile or chaotic training behaviour.

Each architecture is trained for 100 epochs under two conditions: (1) a fixed batch size of 64, establishing a baseline stability profile, and (2) the DEBA regime, which adapts batch size dynamically according to gradient statistics. We run three random seeds (2, 42, 199) to estimate both the mean stability $\mu_S$ and seed sensitivity $\sigma_S$. DEBA-specific metrics such as decision aggressiveness, i.e the rate of batch size change, and convergence stability (standard deviation of loss over time) are also recorded. Architectures are categorized by their baseline $\mu_S$ values and analyzed for consistency between fixed-batch stability and DEBA performance.

\subsection{Results and Analysis}
Table~\ref{tab:stability_taxonomy} summarizes the profiling outcomes, revealing four distinct stability archetypes that correspond closely to DEBA compatibility patterns.

\textbf{Overly Stable Architectures.}  
MobileNet-V3 and ViT-B-16 achieve high baseline stability ($\mu_S>0.54$, $\sigma_S<0.03$), yet their adaptive responses diverge sharply. MobileNet-V3 gains +5.3\% accuracy as DEBA's negative decision aggressiveness (-0.39) triggers frequent rollbacks, intentionally destabilizing the model ($\mu_S$: 0.590→0.258) to escape suboptimal minima. Conversely, ViT-B-16's extremely low gradient variance ($\sigma^2_g \approx 5\times10^{-7}$) and smooth attention-based landscape leave minimal room for improvement (+0.2\%), as its near-deterministic optimization path remains largely unaffected by batch size changes.

\textbf{Moderately Stable Architectures.}  
ResNet-18 and DenseNet-121 fall within the optimal stability range ($0.48 \leq \mu_S \leq 0.50$), achieving consistent gains (2–3\% accuracy, 36–43\% speedup). DEBA reduces gradient variance by 69–85\% and gradient norm variation by 73\%, smoothing convergence without eliminating stochasticity. Their high decision aggressiveness (0.52–0.7) demonstrates tolerance to rapid batch scaling, facilitated by residual/dense connections that preserve gradient flow. These results validate the "sweet spot" hypothesis, in which DEBA reinforces convergence efficiency without destabilization.

\textbf{Dynamically Stable Architecture}  
EfficientNet-B0 exhibits intermediate stability ($\mu_S=0.331$) but extreme seed sensitivity ($\sigma_S=0.182$). DEBA consistently improves accuracy by 1.5–2.4\% through responsive rollback mechanisms (aggressiveness -0.02 to -0.12) that counteract volatility, demonstrating intelligent navigation of chaotic dynamics rather than pure stabilization.

\textbf{Naturally Unstable Architecture}  
ResNet-50 ($\mu_S=0.185$, $\sigma_S=0.079$) exhibits highly seed-dependent outcomes: favourable initializations yield +2.17\% accuracy, while unstable ones cause -1.6\% degradation. Its depth amplifies gradient interactions, causing DEBA's aggressiveness (0.28–0.4) to exacerbate rather than correct instability, confirming that naturally unstable architectures are unreliable candidates for adaptive scheduling.

\subsubsection{Predictive Framework}
Baseline stability scores reliably predict DEBA outcomes: architectures within $\mu_S \in [0.45, 0.55]$ consistently benefit; those above 0.55 improve only with convergence stability $<$ 0.20 (indicating stagnation); models below 0.26 show seed-dependent behaviour. EfficientNet-B0 is correctly identified as "dynamically stable" via $\sigma_S>0.15$. Decision aggressiveness metrics further confirm DEBA's implicit architecture detection: stable models tolerate rapid scaling (0.52–0.7), while sensitive ones require caution (0.28–0.45), with negative values (-0.39 to -0.10) indicating active rollback cycles.

\begin{resultbox}
\textbf{Summary:} RQ$_2$ demonstrates that an architecture’s intrinsic training stability provides a strong prior for predicting its response to adaptive batch size scheduling. The stability score $S$ computed from fixed-batch runs captures the essential dynamics that govern DEBA’s effectiveness: moderate-stability architectures benefit consistently, overly stable ones gain from controlled disruption, dynamically stable ones profit from adaptive correction, and naturally unstable ones remain unreliable. This predictive framework offers an alternative to trial-and-error adaptive training, enabling architecture-aware deployment of batch size adaptation strategies.
\end{resultbox}

\begin{table}[ht]
\centering
\caption{Architecture Stability Taxonomy (Fixed BS Profiling)}
\label{tab:stability_taxonomy}
\resizebox{\columnwidth}{!}{%
\begin{tabular}{lcccc}
\hline
\textbf{Architecture} & $\boldsymbol{\mu_S}$ & $\boldsymbol{\sigma_S}$ & \textbf{Class} & \textbf{$\Delta$Acc} \\
\hline
MobileNet-V3 & 0.590 & 0.028 & Overly Stable & +5.3\% \\
ViT-B-16 & 0.546 & 0.012 & Overly Stable & +0.2\% \\
\hline
DenseNet-121 & 0.495 & 0.019 & Moderately Stable & +1.9\% \\
ResNet-18 & 0.487 & 0.015 & Moderately Stable & +2.6\% \\
\hline
EfficientNet-B0 & 0.331 & 0.182 & Dynamically Stable & +1.9\% \\
\hline
ResNet-50 & 0.185 & 0.079 & Naturally Unstable & -0.3\% \\
\hline
\end{tabular}%
}
\end{table}

\section{RQ$_3$: \rqthree}
\label{sec:RQ3}

\subsection{Motivation}

Section~\ref{sec:RQ1} demonstrates that architecture fundamentally determines adaptive batch size effectiveness, with speedup varying from 8\% to 62\% and accuracy changes from -0.6\% to +7.0\%. We investigate whether these outcomes reflect intrinsic architectural properties or scheduler implementation artifacts through systematic ablations of: (1) threshold tuning strategy, (2) stabilization period, and (3) observation window for gradient statistics.

\subsection{Methodology}
Our baseline DEBA uses architecture-specific thresholds from profiling (Table~\ref{tab:thresholds}), cooldown=10 epochs, sliding window=15 epochs, and growth/rollback factors of 1.5×/0.8×, achieving 36-62\% speedup with accuracy gains on 5/6 architectures. All experiments use CIFAR-10, 100 epochs, 5 random seeds, with fixed BS=64 as comparison.

\subsection{Results}
Table~\ref{tab:ablation_combined} reveals three patterns: (1) threshold calibration and cooldown critically affect all architectures, (2) observation window has moderate, architecture-dependent effects, and (3) no single configuration succeeds universally.

\subsubsection{Ablation 1: Threshold Calibration Strategy}
Our baseline derives architecture-specific thresholds through profiling (Table~\ref{tab:thresholds}), revealing a 48× range: $\theta_{\text{stab}}$ spans 0.25 (\mobilenet-V3) to 12.0 (\resnet-50). Universal thresholds ($\theta_{\text{stab}}=0.1$, $\theta_{\text{conf}}=0.5$) catastrophically degrade half the architectures: \resnet-18 loses all benefits ($\Delta$Acc: +2.43→-0.16, speedup: -10.8\%), while \densenet-121 and \efficientnet-B0 drop 1.7-3.3 points with speedup regressions to -25.5\%. The failure reflects gradient scale mismatch—architectures with moderate gradient norm variation (0.4-0.6) find $\theta_{\text{stab}}=0.1$ too restrictive, triggering excessive rollbacks. Conversely, \mobilenet-V3's success (+7.02\%) is coincidental, as its gradient scale (0.2-0.3) happens to match defaults. The 48× threshold range reflects fundamental architectural differences in gradient behaviour, not tuning artifacts.

\subsubsection{Ablation 2: Stabilization Period Between Adaptations}

Cooldown=2 universally degrades performance, with 5/6 architectures losing $>1$ accuracy point. \resnet-50 suffers catastrophic collapse (-12.62 points despite 58.7\% speedup), while even high-benefit \mobilenet-V3 retains only marginal gains (+0.64 vs. +6.98 baseline). Training curves reveal "decision thrashing" where batch size oscillates every 2-3 epochs, preventing convergence.

Cooldown=5 recovers most architectures (+0.24 to +5.24 points, 50-67\% speedup), though \resnet-50 remains problematic (-2.35 points). Cooldown=10 provides incremental accuracy improvements (\resnet-18: +2.43 vs. +0.83; \mobilenet-V3: +6.98 vs. +5.24) at modest speedup cost (\densenet-121: 67\%→55\%), representing a fundamental speed-accuracy tradeoff.

Three processes require stabilization: BatchNormalization recalibration, momentum buffer adjustment, and loss landscape navigation after effective learning rate changes. Cooldown=2 violates all requirements, causing "adaptation shock." Cooldowns $\geq 5$ provide sufficient recovery, with 10 epochs offering additional stability margin.

\subsubsection{Ablation 3: Temporal Scope of Gradient Statistics}

Full history (vs. sliding window=15) produces heterogeneous outcomes. \efficientnet-B0 and \densenet-121 experience severe speedup collapse (52\%→7\%, 67\%→37\%) as stale statistics cause overly conservative decisions. Conversely, \resnet-18 and \resnet-50 show slight accuracy improvements (+1.94, +0.14) with reduced speedup—full history's conservatism prevents harmful scaling but sacrifices efficiency. \mobilenet-V3 and ViT-B16 remain unaffected due to stable, slowly-varying gradient distributions.

The failure stems from two mechanisms: (1) stale reference statistics inflate baselines, triggering false rollback alarms, and (2) adaptation lag where historical averages dilute recent signals from qualitatively different training phases. Sliding windows detect regime transitions within 15 epochs, enabling responsive adaptation.

\subsection{Design Principles for Robust Adaptive Scheduling}

\textbf{Universal requirements.} Sufficient stabilization period (cooldown $\geq 5$ epochs) is architecture-invariant. Cooldown=2 causes 2.3-12.6 point losses across all models due to optimization fundamentals: BatchNorm statistics, momentum buffers, and loss landscape trajectories require 3-5 epochs to equilibrate. We recommend a minimum cooldown of 5, increasing to 10 for high-seed-sensitivity models. % or when prioritizing accuracy.

Sliding windows provide substantially more consistent behaviour than full history, balancing responsiveness with statistical stability. Full history causes unpredictable outcomes: speedup variance explosion (\efficientnet-B0: std 6\%→53\%) and average speedup collapse (\densenet: 67\%→37\%).

\textbf{Architecture-specific requirements.} Threshold calibration is non-negotiable. Universal thresholds degrade 50\% of architectures (3/6) by 1.7-3.3 points with speedup regressions to -25.5\%. The 48× variation in optimal $\theta_{\text{stab}}$ reflects fundamental gradient scale differences. We recommend baseline profiling (Section~\ref{sec:RQ2})—a single 100-epoch fixed-batch run —to derive architecture-specific thresholds.

\begin{resultbox}
\textbf{Summary:} RQ$_3$ reveals that implementation details are not created equal. Cooldown period is a hard constraint requiring minimum values regardless of architecture, while threshold calibration is essential but architecture-dependent. Combined with baseline profiling (Section \ref{sec:RQ2}), these principles provide a structured deployment procedure grounded in measurable gradient behaviour, reducing brittleness in adaptive methods.
\end{resultbox}

\begin{table*}[ht]
\centering
\caption{Impact of implementation choices on DEBA performance (CIFAR-10, 5 seeds). 
Baseline uses architecture-specific thresholds, cooldown=10, and sliding window=15. 
$\Delta$Acc shows accuracy change vs.\ fixed BS=64 baseline.}
\label{tab:ablation_combined}
\small
\setlength{\tabcolsep}{3.5pt}
\begin{tabular}{lcccccccccc}
\toprule
& \multicolumn{2}{c}{\textbf{Baseline}} 
& \multicolumn{2}{c}{\textbf{Universal Thresh.}} 
& \multicolumn{2}{c}{\textbf{Cooldown=2}} 
& \multicolumn{2}{c}{\textbf{Cooldown=5}} 
& \multicolumn{2}{c}{\textbf{Full History}} \\
\cmidrule(lr){2-3} \cmidrule(lr){4-5} \cmidrule(lr){6-7} \cmidrule(lr){8-9} \cmidrule(lr){10-11}
\textbf{Architecture} 
& $\Delta$Acc & Speedup (\%) 
& $\Delta$Acc & Speedup (\%) 
& $\Delta$Acc & Speedup (\%) 
& $\Delta$Acc & Speedup (\%) 
& $\Delta$Acc & Speedup (\%) \\
\midrule
ResNet-18       & +2.43 & 34.2 & -0.16 & -10.8 & -2.75 & 40.7 & +0.83 & 21.6 & +1.94 & 31.8 \\
ResNet-50       & -0.59 & 42.4 & +0.43 & -5.6  & -12.62 & 58.7 & -2.35 & 43.4 & +0.14 & 11.9 \\
DenseNet-121    & +1.40 & 54.9 & -0.26 & -0.05 & -4.30 & 79.0 & +0.24 & 67.0 & +1.02 & 37.1 \\
EfficientNet-B0 & +2.37 & 43.3 & -0.95 & -25.5 & -2.29 & 65.0 & +1.97 & 52.0 & +0.24 & 7.1 \\
MobileNet-V3    & +6.98 & 50.4 & +7.02 & 37.9  & +0.64 & 55.4 & +5.24 & 50.2 & +5.48 & 42.0 \\
ViT-B16         & -0.61 & 8.3  & -0.02 & -1.9  & -1.80 & 11.9 & -0.59 & 10.0 & -0.43 & 6.8 \\
\midrule
\textit{Degraded ($>$1pt)} & 0/6 & --- & 3/6 & --- & 5/6 & --- & 1/6 & --- & 0/6 & --- \\
\textit{Mean $|\Delta$Acc|} & 2.40 & --- & 1.47 & --- & 4.07 & --- & 1.20 & --- & 1.54 & --- \\
\bottomrule
\end{tabular}
\end{table*}

\section{Discussion}
\label{sec:discussion}

\subsection{Key Insights and Generalization}
Our findings show that the effectiveness of adaptive batch scheduling is not universal but fundamentally determined by architectural characteristics. Lightweight and densely connected models consistently benefit from DEBA, whereas very deep or inherently stable ones, such as ResNet-50 and ViT-B16, exhibit limited or unstable gains. These trends remain consistent across datasets and random seeds, indicating that the determining factor is architectural design rather than random variation. Although our experiments focus on CIFAR-scale data, the mechanisms identified—gradient variance control, exploration–stability balance, and architecture-dependent noise tolerance—are fundamental to stochastic optimization and are expected to generalize to larger-scale training scenarios.

\subsection{Practical Implications}
DEBA’s profiling-based configuration provides a practical route toward architecture-aware adaptation. A fixed-batch profiling run can estimate gradient stability and guide the selection of architecture-specific thresholds, while a moderate cooldown between adaptations helps ensure stability. This lightweight procedure reliably reproduces DEBA’s benefits and mitigates instability across architectures without the need for costly hyperparameter searches.

\subsection{Limitations and Interactions}
The proposed taxonomy, which distinguishes high-, moderate-, and low-benefit architectures, is empirical and may not directly apply to extremely deep or hybrid designs such as ConvNeXt or CoAtNet. Furthermore, our controlled setup isolates batch-size adaptation while keeping the learning rate fixed. In contrast, in realistic training pipelines, DEBA would interact with adaptive learning rate schedules, warmup phases, and normalization layers. Understanding these interactions and extending DEBA to co-adapt batch size and learning rate, as in AdaScale-like frameworks, represents an essential next step.

\subsection{Future Directions}
Future work will extend the evaluation of DEBA to larger-scale datasets such as ImageNet and to distributed training environments where profiling overhead becomes more pronounced. We also plan to explore meta-learning or transfer-based approaches that can infer optimal adaptation thresholds directly from architectural features, reducing manual calibration. Applying this framework to other domains, including natural language processing and reinforcement learning, would further test whether architecture-dependent stability generalizes across modalities. Finally, establishing formal connections between gradient statistics, Lipschitz smoothness, and adaptive behavior remains an important direction toward a theoretical foundation for architecture-aware optimization.

\section{Threats to Validity}
\label{sec:threats}

\subsection{Internal validity} We fix the learning rate (0.01) and optimizer (SGD, momentum=0.9) to isolate batch size effects. While large-batch training often uses LR scaling~\cite{goyal2017accurate} or adaptive optimizers, our controlled setup ensures observed architecture-dependent patterns reflect intrinsic optimization characteristics rather than optimizer interactions. DEBA's hyperparameters (thresholds, cooldown) are tuned via baseline profiling (Section~\ref{sec:RQ2}). Our ablations (Section~\ref{sec:RQ3}) vary factors independently; subtle interactions may exist, but the 48× threshold range confirms architecture-agnostic tuning is fundamentally inadequate.

\subsection{External validity} We evaluate on CIFAR-10/100 (32×32, 50k samples), enabling rigorous multi-seed evaluation but potentially understating large-scale dynamics (ImageNet: 1.2M samples, 224×224). Architecture-dependent trends remain consistent across both datasets (10 vs 100 classes), suggesting these are not dataset artifacts. Our six architectures span diverse paradigms (residual, dense, efficient, and attention-based) but do not cover all modern families (such as ConvNeXt and Swin). The baseline profiling method (Section~\ref{sec:RQ2}) enables assessing new architectures without complete retraining. Single-GPU experiments (H100) offer maximal control; distributed settings may alter speedups due to communication overhead, though per-epoch time reductions stem from hardware-agnostic parallelism principles.

\subsection{Statistical validity} We use five random seeds per configuration (60 runs total), exceeding norms in adaptive batch studies~\cite{devarakonda2017adabatch,smith2018dontdecaylearningrate}. Our strongest claims rely on consistent, low-variance gains (std ${<}1.5\%$); high variance in problematic models (\resnet-50) is treated as an empirical finding. Despite fixed seeds, GPU-level randomness (cuDNN) may cause minor deviations~\cite{pineau2020improvingreproducibilitymachinelearning}. Consistent trends across diverse seeds (2, 7, 42, 123, 199) demonstrate robustness.

\section{Conclusion}
\label{sec:conclusion}

This work challenges the assumption that a single adaptive batch size strategy can generalize across all neural network architectures. We introduced \textbf{DEBA}, a multi-signal scheduler that dynamically adjusts batch size based on gradient and loss dynamics. Through systematic evaluation across six architectures and two datasets, we showed that architecture fundamentally determines the effectiveness of adaptive batch scheduling. Lightweight and moderately deep networks consistently benefit, while very deep or intrinsically stable ones show limited or even negative responses, revealing three distinct regimes of architectural behaviour.

Beyond these empirical findings, DEBA establishes a practical framework for architecture-aware adaptation. Its gradient-based signals offer interpretable criteria for batch size scaling, and its baseline profiling procedure allows practitioners to anticipate compatibility through short fixed-batch runs. This approach replaces heuristic tuning with predictive insight, enabling selective and efficient deployment of adaptive methods.

More broadly, our results indicate that adaptation in deep learning cannot remain architecture-agnostic. As models continue to diversify—from compact edge networks to billion-parameter transformers—training strategies must evolve toward systems that recognize and respond to architectural properties. DEBA thus represents both a methodological advance and a conceptual step toward principled, architecture-aware optimization. All code and experimental logs will be released in our replication package~\cite{anonymous_2025_17478685} to support reproducibility and future research.

\nocite{langley00}
\balance
\bibliography{example_paper}
\bibliographystyle{mlsys2025}
%\newpage
\appendix
\section{Appendix}

\begin{table*}[ht]
\centering
\caption{Training performance with fixed batch size (64) vs. DEBA across six architectures on CIFAR-10 and CIFAR-100. Mean and standard deviation over five random seeds. \textcolor{green}{Green} indicates improvement, \textcolor{red}{red} indicates degradation.}
\label{tab:main_results}
\resizebox{\textwidth}{!}{
\begin{tabular}{llccccccc}
\toprule
\textbf{Architecture} & \textbf{Dataset} & \textbf{Method} & \textbf{Accuracy (\%)} & \textbf{$\Delta$ Acc} & \textbf{Time (s)} & \textbf{Speedup (\%)} & \textbf{Throughput} \\
\midrule
\multirow{2}{*}{\resnet-18} 
& \multirow{2}{*}{CIFAR-10} 
& Fixed & $83.05 \pm 0.51$ & -- & $778 \pm 38$ & -- & 1.0×\\
& & DEBA & $\mathbf{85.48 \pm 0.66}$ & \textcolor{green}{+2.43} & $512 \pm 101$ & $36.6 \pm 13.1$ & 1.5×\\
\cmidrule{2-8}
& \multirow{2}{*}{CIFAR-100} 
& Fixed & $51.81 \pm 0.56$ & -- & $774 \pm 32$ & -- & 1.0×\\
& & DEBA & $\mathbf{55.79 \pm 0.77}$ & \textcolor{green}{+3.98} & $441 \pm 15$ & $43.0 \pm 2.9$ & 1.8×\\
\midrule
\multirow{2}{*}{\resnet-50} 
& \multirow{2}{*}{CIFAR-10} 
& Fixed & $\mathbf{83.07 \pm 0.62}$ & -- & $1858 \pm 28$ & -- & 1.0×\\
& & DEBA & $82.48 \pm 1.62$ & \textcolor{red}{-0.59} & $1069 \pm 120$ & $43.3 \pm 5.5$ & 1.7×\\
\cmidrule{2-8}
& \multirow{2}{*}{CIFAR-100} 
& Fixed & $50.14 \pm 0.81$ & -- & $1887 \pm 45$ & -- & 1.0×\\
& & DEBA & $\mathbf{51.43 \pm 3.85}$ & \textcolor{green}{+1.29} & $1045 \pm 81$ & $44.7 \pm 3.8$ & 1.8×\\
\midrule
\multirow{2}{*}{\densenet-121} 
& \multirow{2}{*}{CIFAR-10} 
& Fixed & $84.95 \pm 0.57$ & -- & $2906 \pm 73$ & -- & 1.0×\\
& & DEBA & $\mathbf{86.35 \pm 0.68}$ & \textcolor{green}{+1.40} & $1310 \pm 197$ & $55.0 \pm 6.6$ & 2.2×\\
\cmidrule{2-8}
& \multirow{2}{*}{CIFAR-100} 
& Fixed & $56.18 \pm 0.70$ & -- & $2923 \pm 46$ & -- & 1.0×\\
& & DEBA & $\mathbf{59.42 \pm 0.08}$ & \textcolor{green}{+3.24} & $1097 \pm 23$ & $62.4 \pm 1.1$ & 2.7×\\
\midrule
\multirow{2}{*}{\efficientnet-B0} 
& \multirow{2}{*}{CIFAR-10} 
& Fixed & $83.18 \pm 0.65$ & -- & $1443 \pm 61$ & -- & 1.0×\\
& & DEBA & $\mathbf{85.55 \pm 0.63}$ & \textcolor{green}{+2.37} & $818 \pm 210$ & $47.7 \pm 13.1$ & 1.8×\\
\cmidrule{2-8}
& \multirow{2}{*}{CIFAR-100} 
& Fixed & $56.29 \pm 0.52$ & -- & $1434 \pm 56$ & -- & 1.0×\\
& & DEBA & $\mathbf{56.72 \pm 0.85}$ & \textcolor{green}{+0.43} & $618 \pm 7$ & $56.8 \pm 1.5$ & 2.3×\\
\midrule
\multirow{2}{*}{\mobilenet-V3} 
& \multirow{2}{*}{CIFAR-10} 
& Fixed & $69.54 \pm 3.18$ & -- & $882 \pm 45$ & -- & 1.0×\\
& & DEBA & $\mathbf{76.52 \pm 0.45}$ & \textcolor{green}{+6.98} & $438 \pm 9$ & $50.3 \pm 3.4$ & 2.0×\\
\cmidrule{2-8}
& \multirow{2}{*}{CIFAR-100} 
& Fixed & $41.36 \pm 0.76$ & -- & $899 \pm 61$ & -- & 1.0×\\
& & DEBA & $\mathbf{45.37 \pm 1.03}$ & \textcolor{green}{+4.01} & $439 \pm 3$ & $51.0 \pm 3.3$ & 2.0×\\
\midrule
\multirow{2}{*}{\vitb} 
& \multirow{2}{*}{CIFAR-10} 
& Fixed & $\mathbf{68.84 \pm 1.2}$ & -- & $16720.31 \pm 139.3$ & -- & 1.0×\\
& & DEBA & $68.23 \pm 0.56$ & \textcolor{red}{-0.61} & $15331.05 \pm 479.71$ & $8.3 \pm 2.89$ & 1.01×\\
\cmidrule{2-8}
& \multirow{2}{*}{CIFAR-100} 
& Fixed & $38.57 \pm 1.1$ & -- & $16719.13 \pm 293.1$ & -- & 1.0×\\
& & DEBA & $\mathbf{39.62 \pm 2.34}$ & \textcolor{green}{+1.05} & $15859.8 \pm 331.02$ & $5.1 \pm 2.96$ & 1.05×\\
\bottomrule
\end{tabular}}
\end{table*}

\begin{figure*}[ht]
    \centering
    \includegraphics[width=\textwidth]{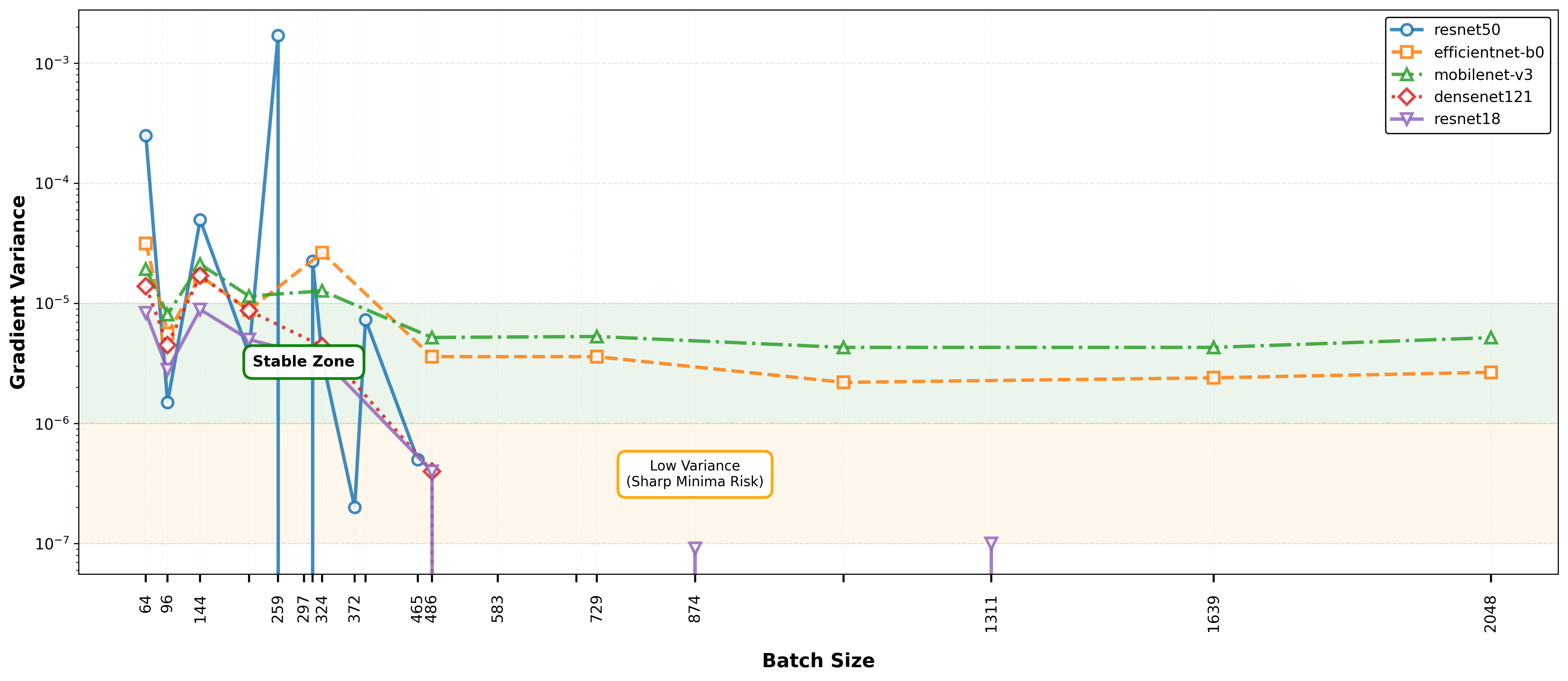}
    \vspace{-2mm}
    \caption{Gradient variance as a function of batch size on CIFAR-100.} %The optimal range (10$^{-6}$ to 10$^{-5}$, shaded) balances convergence stability with implicit regularization. DEBA's adaptive scheduling navigates this landscape dynamically.}
    \label{fig:gradient_Variance_cifar_100}
\end{figure*}

\end{document}